%% file: preamble.tex
\def\papersize{letter}

\documentclass[sigconf,\papersize]{acmart}

\usepackage{epsfig,endnotes}
\usepackage[utf8]{inputenc}

\usepackage{amsmath,amsthm}
\usepackage{bbm}
\usepackage{mathtools}
\usepackage[nounderscore]{syntax}
\usepackage{algorithm,algpseudocode}

\usepackage{import}
\usepackage{graphicx}
\graphicspath{ {figures/} }
\usepackage{subcaption}
\usepackage{stmaryrd}
\usepackage{url}
\usepackage{standalone}
\usepackage{cancel}
\usepackage{listings}
\usepackage{xspace}
\usepackage{extarrows}
\usepackage{tabularx}
\usepackage{array}
\usepackage{xcolor}
\usepackage{tikz}
\usetikzlibrary{
  shapes,calc,arrows,fit,positioning,%
  decorations.pathmorphing,snakes,intersections,%
  shapes.geometric,trees%
}
\usepackage{booktabs}
\usepackage{multirow}
\usepackage{adjustbox}

\input{info/defs}
\input{info/commenting}

\setcopyright{none}
\acmYear{}
\copyrightyear{}
\acmDOI{}
\acmISBN{}

\acmConference[AdvML '21]{AdvML '21: Workshop on Adversarial Learning Methods for Machine Learning and Data Mining}{August 15, 2021}{Singapore (virtual)}
\acmBooktitle{AdvML '21: Workshop on Adversarial Learning Methods for Machine Learning and Data Mining, August 15, 2021, Singapore (virtual)}
\begin{document}

%%%%% TITLE, AUTHOR, OTHERS %%%%%

%%% title
\title{Unsupervised Detection of Adversarial Examples with~Model~Explanations}

%%% authors
\input{info/authors}

%%% abstract
\input{abstract}
\maketitle

%%% sections
% introduction
\input{01-intro}
% background
\input{02-background}
% method
\input{03-method}
% evaluation
\input{04-evaluation}
% evaluation
\input{05-conclusion}

% ack
\input{info/ack}

%%% bibliography
% \printbibliography
\bibliographystyle{ACM-Reference-Format}
\bibliography{%
  bibs/example.bib
}

\newpage
\appendix
\input{appendix}

\end{document}

%% file: info/defs.tex
\makeatletter
\def\thm@space@setup{\thm@preskip=1pt
\thm@postskip=1pt}
\makeatother

\newcommand{\code}[1]{\lstinline!#1!}

\definecolor{mygray}{rgb}{0.5,0.5,0.5}
\lstdefinestyle{customlisp}{
  belowcaptionskip=1\baselineskip,
  breaklines=true,
  language=Lisp,
  showstringspaces=false,
  numbers=left,
  xleftmargin=2em,
  framexleftmargin=1.5em,
  numbersep=5pt,
  numberstyle=\tiny\color{mygray},
  basicstyle=\small\ttfamily,
  keywordstyle=\color{blue},
  commentstyle=\itshape\color{purple!40!black},
  stringstyle=\color{orange},
  morekeywords={def-rule, ?},
  tabsize=2
}

\makeatletter
\renewcommand{\paragraph}{%
  \@startsection{paragraph}{4}%
  {\z@}{0.75ex \@plus 0.5ex \@minus .2ex}{-1em}%
  {\normalfont\normalsize\bfseries}%
}
\makeatother

\usepackage{enumitem}
\setlist{nosep}

\algtext*{EndFor}
\algtext*{EndIf}
\algtext*{EndProcedure}

%% file: info/commenting.tex
\usepackage{soul}
\definecolor{lightgreen}{rgb}{0.85,1.00,0.85}
\definecolor{lightorange}{rgb}{1.00,0.90,0.50}
\definecolor{lightred}{rgb}{1.00,0.85,0.85}
\definecolor{lightblue}{rgb}{0.85,0.85,1.00}
\definecolor{lightpurple}{rgb}{1.00,0.85,1.00}
\DeclareRobustCommand{\commentformat}[3]{\sethlcolor{#2}\hl{\textsf{#1: #3}}}

\newcommand{\cut}[1]{}

\ifdefined\draft
\else
\renewcommand{\commentformat}[3]{}
\fi

%% file: info/authors.tex
%!TEX root=preamble.tex

%%% AUTHOR INFORMATION %%%
\author{Gihyuk Ko}
\authornote{Work done at CSRC (CyberSecurity Research Center) in KAIST.}
\affiliation{
    \institution{Carnegie Mellon University}
    \city{Pittsburgh}
    \country{USA}
}
\affiliation{
    \institution{CSRC, KAIST}
    \city{Daejeon}
    \country{South Korea}
}
\email{gihyuk.ko@gmail.com}

\author{Gyumin Lim}
\affiliation{
    \institution{CSRC, KAIST}
    \city{Daejeon}
    \country{South Korea}
}
\email{6sephiruth@kaist.ac.kr}

%% file: abstract.tex
%!TEX root=preamble.tex

%%% ABSTRACT %%%
\begin{abstract}
Deep Neural Networks (DNNs) have shown remarkable performance in a diverse range of machine learning applications. However, it is widely known that DNNs are vulnerable to simple adversarial perturbations, which causes the model to incorrectly classify inputs. In this paper, we propose a simple yet effective method to detect adversarial examples, using methods developed to \emph{explain} the model's behavior. Our key observation is that adding small, humanly imperceptible perturbations can lead to drastic changes in the model explanations, resulting in \emph{unusual} or \emph{irregular} forms of explanations. From this insight, we propose an unsupervised detection of adversarial examples using reconstructor networks trained only on model explanations of benign examples. Our evaluations with MNIST handwritten dataset show that our method is capable of detecting adversarial examples generated by the state-of-the-art algorithms with high confidence. To the best of our knowledge, this work is the first in suggesting unsupervised defense method using model explanations.
\end{abstract}

%% file: 01-intro.tex
%!TEX root=preamble.tex

\section{Introduction}
\label{sec:intro}

Deep neural networks have shown remarkable performance in complex real-world tasks including image and audio classification, text recognition and medical applications. However, they are known to be vulnerable to adversarial examples -- adversarially perturbed inputs which can be easily generated to fool the decisions made by DNNs~\cite{szegedy13ae,eykholt18ae}. Such attacks can lead to devastating consequences, as they can undermine the security of the system deep networks are being used.

In order to prevent such attacks from happening, many recent efforts have focused on developing methods in detecting adversarial examples~\cite{gong17,grosse17,tian18,fidel20} and preventing their usage. However, many existing works suffer from high computational cost, because they rely on pre-generated adversarial examples.

In this work, we suggest a simple yet effective method in detecting adversarial examples; our method uses model explanations in an \emph{unsupervised} manner, meaning that no pre-generated adversarial samples are required. Our work motivates from the insight that a small perturbation to the input can result in large difference in model's explanations. We summarize our contributions as follows:
\begin{itemize}
\item We propose a novel method in detecting adversarial examples, using model explanations. Unlike many previous attempts, our method is \emph{attack-agnostic} and does not rely on pre-generated adversarial samples.
\item We evaluate our method using MNIST, a popular handwritten digit dataset. The experimental results show that our method is comparable to, and often outperforms existing detection methods.
\end{itemize}

%% file: 02-background.tex
%!TEX root=preamble.tex

\section{Background}
\label{sec:background}

In this section, we provide a brief overview on a number of adversarial attacks as well as model explanation used in our experiments. We also briefly discuss on the existing approaches in detection of adversarial examples.

\subsection{Adversarial Examples}

\subsubsection{Fast Gradient Sign Method (FGSM)}
Goodfellow et al. \cite{FGSM_paper} suggested Fast Gradient Sign Method (FGSM) of crafting adversarial examples, which takes the gradient of the loss function with respect to a given input and adds perturbation as a step of size $\epsilon$ in the direction that maximizes the loss function. Formally, for a given parameter $\epsilon$, loss function $\mathcal{L}$, and model parameters $\theta$, input $x$, and label $y$, adversarial example $x'$ is computed as follows:
\begin{align*}
x' = x + \epsilon \cdot \text{sgn} \left[ \nabla_x \mathcal{L}(\theta;x,y) \right],
\end{align*}
where $\text{sgn}\left[\cdot\right]$ is a sign function.

\subsubsection{Projected Gradient Descent (PGD)}
Projected Gradient Descent (PGD)~\cite{PGD_paper} is a multi-step, iterative variant of FGSM which maximizes the cost function via solving following equation:
\begin{align*}
x'_{t+1} = \mathrm{\Pi}_{x+S} \Big( x'_t + \epsilon \cdot \text{sgn} \left[ \nabla_x \mathcal{L}(\theta;x,y) \right] \Big),
\end{align*}
where $x'_t$ is the adversarial example at the step $t$, $\Pi$ is the projection onto the ball of the maximum possible perturbation $x+S$. Solving the optimization over multiple iterations makes PGD more efficient than FGSM, resulting in a more powerful first-order adversary.

\subsubsection{Momentum Iterative Method (MIM)}
Momentum Iterative Method (MIM)~\cite{MI-FGSM_paper} is another variant of FGSM, where it uses gradient velocity vector to accelerate the updates. Adversarial example $x'$ can be obtained from $x$ by solving the following constrained optimization problem:
\begin{align*}
g_{t+1} = \mu \cdot g_t + \frac{ \nabla_x L(x'_t,y) } {\lVert \nabla_x L(x'_t,y) \rVert } \\
x'_{t+1} = x'_t + \epsilon \cdot \text{sgn} \left[ g_{t+1} \right]
\end{align*}
Here, $g_t,x'_t$ represents the value of gradient and generated adversarial example at the step $t$, respectively.

\subsection{Model Explanations}

Due to ever-increasing complexity of deep networks, numerous methods have been developed in order to explain the neural network's behavior. \emph{Input feature attribution} methods are the most widely studied, where they generate local explanations by assigning an \emph{attribution} score to each input feature. Formally, given an input $x = (x_1, \ldots, x_n)$ to a network $f$, feature attribution methods compute $\phi(x,f) := (\phi_1, \ldots, \phi_n) \in \mathbb{R}^n$, assigning score $\phi_i$ to input feature $x_i$.

\paragraph{Input gradient (saliency map).}

One of the first proposed measure of attribution is input gradient~\cite{simonyan13gradient}. Intuitively for a linear function, input gradients represent exact amount that each input feature contributes to the linear function's output. For image inputs, each pixel's contribution could be represented in a heatmap called \emph{saliency map}.

As most practical deep networks compute a \emph{confidence score} for each class label and output the class of with the largest score, multiple saliency maps can be obtained according to the target class label $c$. For simplicity, we only consider the saliency map corresponding to the output class label of the given input. Formally, given an input $x$ and DNN $f$, saliency map of input $x$ is computed as follows:
\begin{align*}
\phi(x,f) := \frac{\partial S_{f(x)}}{\partial x},
\end{align*}
where $S_c$ denotes a confidence score for class label $c$ (i.e., $f(x) := \arg\max_c S_c(x)$).

\subsection{Detection of Adversarial Examples}

Detection-based defenses have been gaining a lot of attention as a potential solution against adversarial attacks. Many works use a supervised approach to train a separate detection neural networks~\cite{gong17,metzen17}, or modify existing network to detect incoming adversarial examples~\cite{grosse17,bhagoji17,li17}. However, these methods often require a large amount of computational cost, where some of them resulting in the loss of accuracy on normal examples~\cite{pang18,tian18}.

Other works apply transformations to the input and analyze (in)consistencies in the outputs of transformed and original inputs. \cite{tian18} uses rotation-based transformation, while \cite{nesti21} suggests a wider variety of transformations such as blurring and adding random noises. While these methods use less computational power, transformations may not be universally applied, and only work for a given dataset.

Similar to our work, \cite{fidel20} trains a classifier separating SHAP~\cite{lundberg17shap} signatures of normal and adversarial examples. However, their method relies on pre-generated adversarial examples, resulting in degraded performance against unknown attacks. Moreover, they use SHAP signatures for the entire class labels instead of a single class, resulting in a large dimension for model explanations as well as high computational cost.

%% file: 03-method.tex
%!TEX root=preamble.tex

\section{Proposed Method}
\label{sec:method}

\begin{figure}[t]
    \begin{center}
        \includegraphics[width=.99\linewidth]{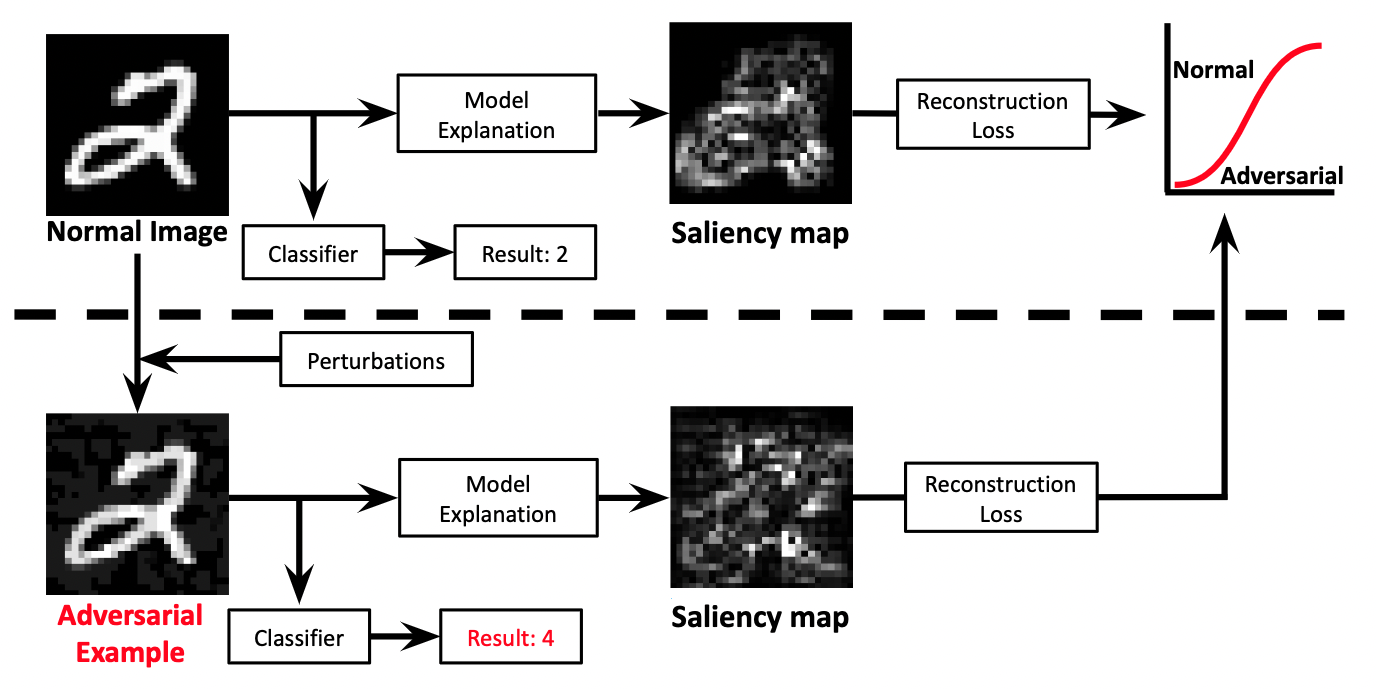}
        \caption{Overview of the proposed detection method.}
        \label{fig:overview}
    \end{center}
\end{figure}

In this section, we illustrate our method: \emph{Unsupervised Detection of Adversarial Examples with Model Explanations}. We first explain the threat model, and then illustrate our approach in detail. An overview of our method is illustrated in Figure~\ref{fig:overview}.

\subsection{Threat Model}
In this paper, we consider an \emph{inspector} for the given machine learning classifier $f$, who wishes to detect (and possibly filter) whether a given input to the model is maliciously crafted to fool the decisions (i.e., the input is an adversarial example). Throughout the paper, we will refer to the model subject to attack as the \emph{target classifier}.

The attacker maliciously crafts adversarial examples in order to fool the decision of the target classifier. We assume that the attacker uses state-of-the-art methods such as FGSM~\cite{FGSM_paper}, PGD~\cite{PGD_paper}, or MIM~\cite{MI-FGSM_paper}, and has access to the training and test samples, as well as the model parameters necessary to conduct the attacks.

\subsection{Our Detection Method}
As noted in Section~\ref{sec:intro}, our method is based on the insight that adding small perterbations to generate adversarial examples could result in \emph{unusual} explanations. Throughout the paper, we denote the explanation of $x$ for DNN $f$ as $\phi(x,f)$. We will often denote it as $\phi(x)$, when $f$ is clear from the context.

Taking advantage of this insight, our method performs unsupervised detection based on three steps: \emph{i) generating input explanations}, \emph{ii) training reconstructor networks using generated explanations}, and \emph{iii) utilizing reconstructor networks to separate normal and adversarial examples.}

\paragraph{Generating input explanations.}
In our proposed method, the inspector is assumed to have access to the training samples $\mathcal{D}_{train}$ that was used to train the target classifier. In order to perform unsupervised anomaly detection based on the model explanations, the inspector first generates \emph{input explanations} for the target model, using training samples.

As noted in Section~\ref{sec:background}, explanations of the target classifier depends on the output label $c$. As the explanations are differently applied for each label, the inspector organizes generated explanations according to the corresponding input's output label. We denote by $\Phi^c$ as a set of input explanations for the inputs in the training dataset with output label $c$.
\begin{align*}
\displaystyle \Phi^c := \Big\{ \phi(x) \  \Big| \  f(x) = c, x \in \mathcal{D}_{train} \Big\}
\end{align*}

\paragraph{Training reconstructor networks.}
Once the explanations for training samples are collected and organized, the inspector then trains \emph{reconstructor networks} --- one for each class label --- which reconstructs the explanations for an input with corresponding class label. Such reconstructor networks will later be used in separating adversarial examples from benign examples.

For each class label $c$, the inspector trains a parameterized network $g(\theta^c;\cdot)$ reconstructing input explanations, solving the following optimization:
\begin{align*}
\theta^c := \arg\min_{\theta} \mathcal{L}_{\Phi^c} \left(\theta; \phi \right) ,
\end{align*}
where $\mathcal{L}_{\Phi}(\theta; \cdot)$ is a reconstruction loss for network $g(\theta; \cdot)$ on $\Phi$.

\paragraph{Separating adversarial examples.}
Lastly, the inspector utilizes the trained reconstructor networks in order to separate adversarial examples from benign examples. As the networks are optimized to reconstruct model explanations of training samples, it will show poor reconstruction quality when an \emph{unusual} shape of explanation is given. Hence, when the reconstruction error is above certain threshold, it is likely that the given input is adversarially crafted.

Formally, for a given suspicious input $x'$, the inspector first obtains the class label $c':=f(x')$ and its explanation $\phi':= \phi(x',f)$. If the reconstruction error of $\phi'$ is larger than given threshold $t_c'$ for label $c'$ (i.e., $\mathcal{L}(\theta^{c'};\phi') > t_c'$), the inspector concludes that the input $x'$ is likely to be an adversarial example.

%% file: 04-evaluation.tex
%!TEX root=preamble.tex

\section{Evaluation}
\label{sec:eval}

In this section, we evaluate the effectiveness of our proposed detection method.

\input{figures/fig-qual}

\subsection{Experimental Setup}

We evaluate our method using the MNIST handwritten digit dataset (MNIST)~\cite{lecun98mnist}. Using MNIST dataset, we first train the \emph{target classifier}, which is subject to the adversarial attacks. In our evaluations, we trained a simple Convolutional Neural Network using the standard 60,000-10,000 train-test split of MNIST dataset. Trained target classifier had >99\% and >98\% classification accuracies for training and test dataset, respectively.

Given the target classifier and the training dataset, model explanations are collected to train a network reconstructing them. In our evaluations, we used \emph{input gradients}~\cite{simonyan13gradient} as model explanations to generate saliency maps. For each class label, the saliency maps for each MNIST training data with corresponding label is collected and used to train the reconstructor network. For all reconstructor networks, we used a simple autoencoder consisting of a single hidden layer. Summary on the model architectures can be found in Table~\ref{tab:arch}.

In order to evaluate the effectiveness of our detection method, we crafted adversarial examples using all 70,000 MNIST images and filtered out unsuccessful attacks (i.e., adding perturbation does not change the original class label). For (successful) adversarial examples, saliency maps were obtained and combined with the saliency maps of the (benign) MNIST test dataset to form a evaluation dataset for our detection method. For a detailed configuration on datasets, we refer to Appendix~\ref{sec:data}.

\input{figures/tab-arch}
\input{figures/tab-comp}

\subsection{Experimental Results}

\paragraph{Effect of input perturbations on explanations.}

Figure~\ref{fig:qual} shows pairs of input image (Input), explanation (Gradient) obtained from the target classifier, and the reconstruction (Recons.) from the trained reconstructor networks, for an example MNIST image and adversarial examples crafted from the image. Here, we confirm our insight that small adversarial perturbations to inputs can lead to noticeable changes in their explanations. Since the reconstructor networks are only trained on benign explanations (input explanations of benign examples), we see that reconstructions of adversarial explanations (input explanations of adversarial examples) are more noisy than the reconstruction of the explanation of original image.

\paragraph{Adversarial detection performance.}

\input{figures/fig-auc}
\input{figures/fig-roc}

In order to evaluate the effectiveness of our proposed detection method, we measure Area Under the ROC Curve (AUC). As our method uses multiple reconstructor networks, we record multiple values of AUC --- each corresponding to a given class label --- as well as their average.

Figure~\ref{fig:auc} shows the trend of AUC values under different adversarial attack scenarios. For each attack, we plot the min, max, and average values of AUC according to the severity of attack (i.e., value of $\epsilon$). While our method has harder time separating adversarial examples with smaller noise level, average AUC stays relatively high even in its lowest values (>95\% for FGSM, >97\% for PGD and MIM, when $\epsilon=0.05$).

Our methods show high performance (average AUC of 0.9583 for FGSM, 0.9942 for PGD, 0.9944 for MIM) in the standard adversarial attack setting of $\epsilon=0.1$ for MNIST dataset. This can be also confirmed by observing Figure~\ref{fig:roc}, where the ROC curves of our detection method for different class labels are plotted.

Note that only a single set (i.e., one per class label) of reconstructor networks is used in all of the attack scenarios. The results clearly show that our defense method can generalize across multiple attack algorithms -- which is impressive, considering that no single adversarial example is provided in the training phase. For a more detailed benchmark results, we refer to Appendix~\ref{sec:bench}.

\paragraph{Quantitative comparison to previous approaches.}

We quantitatively compare our method's adversarial detection accuracy to a number of previous approaches. Specifically, we compare our results with the results from four different existing works (\cite{compare_1,compare_2,compare_3,compare_4}), where the benchmark results are recorded in~\cite{sutanto21}.

Table~\ref{tab:comp} shows comparison on adversarial detection accuracies of the proposed and existing approaches. In all experiments, our method performs the best or the second best in detecting adversarial samples. The results show that our method is comparable to, and often outperforms existing methods.

%% file: figures/fig-qual.tex
%!TEX root=04-evaluation.tex

\newcolumntype{C}{>{\centering\arraybackslash}m{4em}}

\begin{figure}
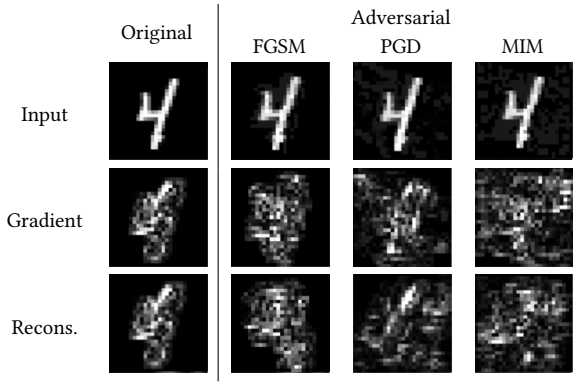

\centering
\begin{tabular}{cC|CCC}
& \multirow{2}{*}{\small Original} & \multicolumn{3}{c}{\small Adversarial} \\
&  & \small FGSM & \small PGD & \small MIM \\
\small Input & \adjustimage{height=1.3cm, valign=1.3cm}{normal_original} & \adjustimage{height=1.3cm, valign=1.3cm}{FGSM_original} & \adjustimage{height=1.3cm, valign=1.3cm}{PGD_original} & \adjustimage{height=1.3cm, valign=1.3cm}{MIM_original} \\
\small Gradient & \adjustimage{height=1.3cm, valign=1.3cm}{normal_saliency} & \adjustimage{height=1.3cm, valign=1.3cm}{FGSM_saliency} & \adjustimage{height=1.3cm, valign=1.3cm}{PGD_saliency} & \adjustimage{height=1.3cm, valign=1.3cm}{MIM_saliency} \\
\small Recons. & \adjustimage{height=1.3cm, valign=1.3cm}{normal_recon} & \adjustimage{height=1.3cm, valign=1.3cm}{FGSM_recon} & \adjustimage{height=1.3cm, valign=1.3cm}{PGD_recon} & \adjustimage{height=1.3cm, valign=1.3cm}{MIM_recon}
\end{tabular}
\caption{Input, gradient, and reconstruction (Recons.) of an example MNIST image and adversarial examples crafted using the image. For each attack, adversarial example with $\epsilon=0.1$ is created.}
\label{fig:qual}
\end{figure}

%% file: figures/tab-arch.tex
%!TEX root=04-evaluation.tex

\begin{table}[t]
\caption{Architectures for target classifier and reconstructor networks.}
\begin{tabular}{cc|cc}
  \hline
  \multicolumn{2}{c|}{\textbf{Target classifier}} & \multicolumn{2}{c}{\textbf{Reconstructor}}\\
  \hline
  Conv.ReLU & 3 $\times$ 3 $\times$ 32 & Dense.ReLU & 784 \\
  Dense.ReLU & 128 & Dense.ReLU & 64  \\
  Softmax & 10 & Dense.ReLU & 784 \\
  \hline
\end{tabular}
\label{tab:arch}
\end{table}

%% file: figures/tab-comp.tex
%!TEX root=04-evaluation.tex

\begin{table}[t]
\caption{Comparison on adversarial detection accuracy of the proposed (Ours) and existing approaches. The best and the second best results are highlighted in boldface and underlined texts, respectively. All benchmarks are done on MNIST dataset.}
\begin{tabular}{lc|cccc|c}
    \multicolumn{2}{c|}{\textbf{Adv. Attack}} & \cite{compare_1} & \cite{compare_2} & \cite{compare_3} & \cite{compare_4} & \textbf{Ours} \\
    \hline \hline
    \multirow{3}{0.7cm}{FGSM} & $\epsilon=0.1$ & 0.7768 & 0.7952 & \textbf{0.9514} & 0.8030 & \underline{0.9233} \\
    & $\epsilon=0.2$ & 0.8672 & 0.8977 & \textbf{0.9826} & 0.7767 & \underline{0.9573} \\
    & $\epsilon=0.3$ & 0.8925 & 0.9380 & \textbf{0.9887} & 0.8681 & \underline{0.9693} \\
    \hline
    \multirow{3}{0.7cm}{PGD (BIM\textsuperscript{*})} & $\epsilon=0.1$ & 0.9419 & 0.8096 & \underline{0.9716} & 0.8092 & \textbf{0.9839}\\
    & $\epsilon=0.2$ & 0.9768 & 0.8330 & \underline{0.9890} & 0.9027 & \textbf{0.9894}\\
    & $\epsilon=0.3$ & 0.9801 & 0.7088 & \underline{0.9896} & 0.9574 & \textbf{0.9901}\\
\end{tabular}
\begin{flushleft}
    \scriptsize
    * BIM (Basic Iterative Method)~\cite{I-FGSM_paper} is a variant of FGSM, similar to PGD. As in some papers PGD is described as a specialized BIM attack, we included results of BIM attacks in the table.
\end{flushleft}
\label{tab:comp}
\end{table}

%% file: figures/fig-auc.tex
%!TEX root=04-evaluation.tex

\begin{figure*}[t]
    \centering
    \begin{subfigure}[b]{0.33\textwidth}
        \centering
        \includegraphics[width=\textwidth]{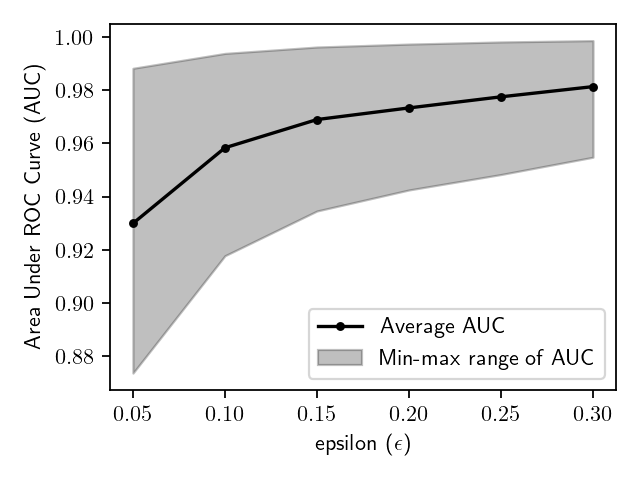}
        \caption{FGSM.}
        \label{fig:auc-fgsm}
    \end{subfigure}
    \hfill
    \begin{subfigure}[b]{0.33\textwidth}
        \centering
        \includegraphics[width=\textwidth]{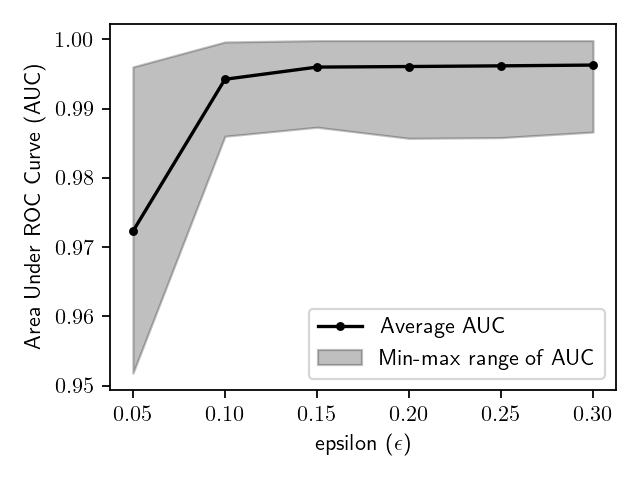}
        \caption{PGD.}
        \label{fig:auc-pgd}
    \end{subfigure}
    \hfill
    \begin{subfigure}[b]{0.33\textwidth}
        \centering
        \includegraphics[width=\textwidth]{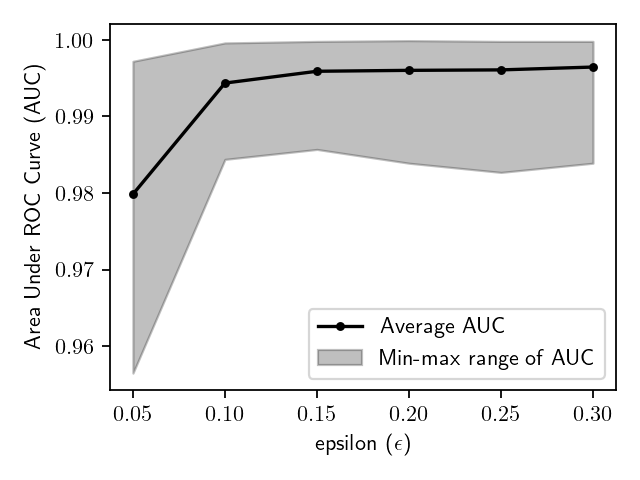}
        \caption{MIM.}
        \label{fig:auc-mim}
    \end{subfigure}
    \caption{Area under the Receiver Operating Characteristic (ROC) curve obtained according to the attack's severity (parameterized by $\epsilon$), for (a) FGSM, (b) PGD, and (c) MIM attacks. For each class label, our proposed detector's performance is recorded using adversarial examples created using given \emph{(attack, epsilon)} pair. Grey areas show the min-max range of AUC, and black lines show average value of AUC across different class labels. All experiments were done using MNIST dataset.}
    \label{fig:auc}
\end{figure*}

%% file: figures/fig-roc.tex
%!TEX root=04-evaluation.tex

\begin{figure*}[t]
    \centering
    \begin{subfigure}[b]{0.33\textwidth}
        \centering
        \includegraphics[width=\textwidth]{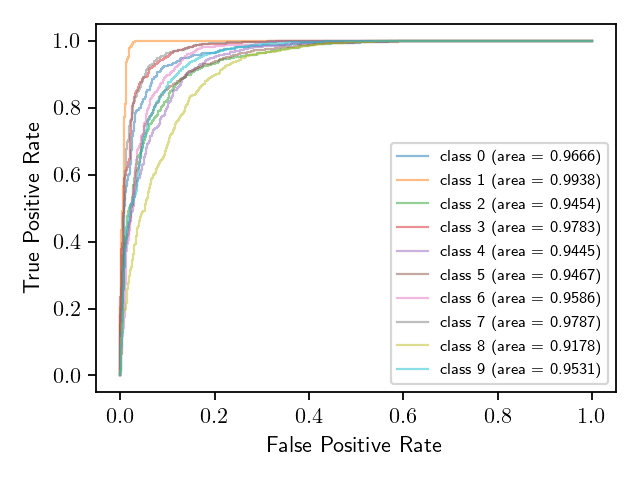}
        \caption{FGSM.}
        \label{fig:roc-fgsm}
    \end{subfigure}
    \hfill
    \begin{subfigure}[b]{0.33\textwidth}
        \centering
        \includegraphics[width=\textwidth]{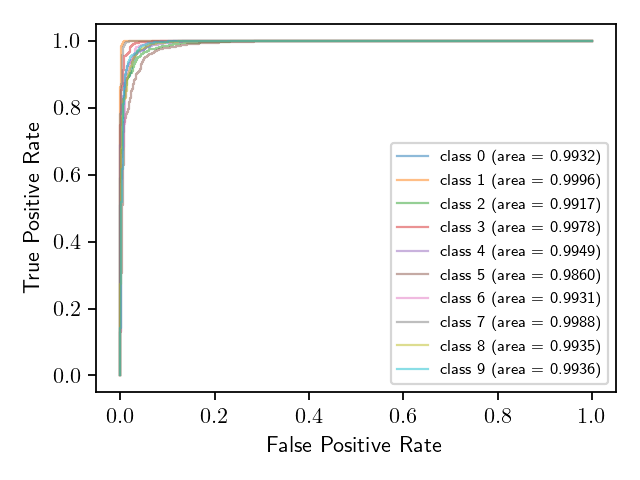}
        \caption{PGD.}
        \label{fig:roc-pgd}
    \end{subfigure}
    \hfill
    \begin{subfigure}[b]{0.33\textwidth}
        \centering
        \includegraphics[width=\textwidth]{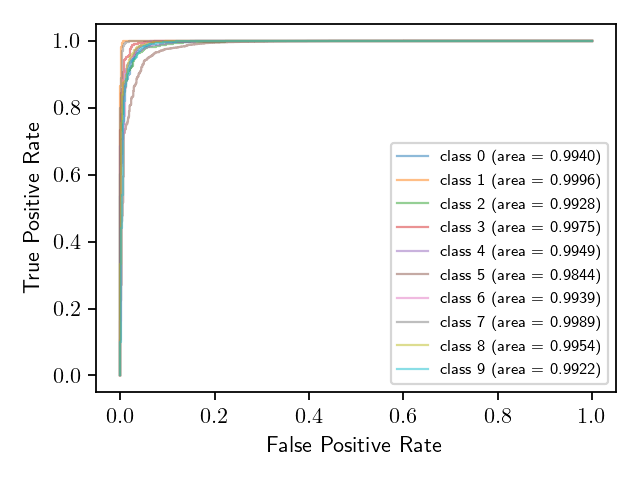}
        \caption{MIM.}
        \label{fig:roc-mim}
    \end{subfigure}
    \caption{Receiver Operating Characteristic (ROC) curve for different class labels, for (a) FGSM, (b) PGD, and (c) MIM attacks of $\epsilon=0.1$. All experiments were done using MNIST dataset.}
    \label{fig:roc}
\end{figure*}

%% file: 05-conclusion.tex
%!TEX root=preamble.tex

\section{Conclusion}
\label{sec:conclusion}

In this paper, we propose a novel methodology in detecting adversarial examples using model explanations. Our method is motivated from the insight that even when small perturbation is added to the input, model explanations can drastically be altered. Taking advantage of this, we suggested an anomaly detection of adversarial examples using a network optimized to reconstruct the model explanations from benign examples. Unlike supervised methods, our method is \emph{attack-agnostic}, in that it does not require pre-generated adversarial samples.

In our experiments using MNIST handwritten dataset, we showed that our method is capable of separating benign and adversarial examples with high performance, comparable to, or better than existing approaches. We argue that our method is more efficient due to its unsupervised manner; with single training of reconstructor networks, multiple state-of-the-art attacks such as FGSM, PGD, and MIM can be prevented. To the best of our knowledge, this work is the first in suggesting unsupervised defense method using model explanations.

%% file: info/ack.tex
%!TEX root=preamble.tex

%%% ACKNOWLEDGEMENTS %%%
\section*{Acknowledgment}
This work was developed with the suppport of Institute of Information \& communications Technology Planning \& Evaluation (IITP) grant, funded by the Korea government (MSIT) (No.2020-0-00153, Penetration Security Testing of ML Model Vulnerabilities and Defense).

%% file: appendix.tex
%!TEX root=preamble.tex

\section{Datasets for Reconstructor Networks}
\label{sec:data}

\input{figures/tab-data}

%Table~\ref{tab:data} contains number of data samples in training and test dataset for reconstructor networks. As explained in Section~\ref{sec:eval}, we craft and filter only successful adversarial examples. While we extensively experiment over many values of epsilons in our evaluations, we only show the dataset numbers by the interval of 0.05.

\section{Detailed Performance Benchmark}
\label{sec:bench}

\input{figures/tab-bench}

%% file: figures/tab-data.tex
%!TEX root=appendix.tex

\begin{table}[H]
\caption{Training and test dataset configurations for training and evaluating reconstructor networks. The number of adversarial samples for each attack scenario are summed up for all class labels.}
\begin{tabular}{ll|c|c|c}
    \multicolumn{2}{c|}{\multirow{2}{*}{\bfseries Adv. Attack}} & \textbf{Training} & \multicolumn{2}{c}{\bfseries Test} \\
    \cline{3-5}
    & & normal & normal & adversarial \\
    \hline \hline
    \multirow{6}{0.7cm}{FGSM} & $\epsilon=0.05$ & \multirow{6}{*}{60000\textsuperscript{*}} & \multirow{6}{*}{10000\textsuperscript{**}} & 5797 \\
    & $\epsilon=0.1$ & & & 22649\\
    & $\epsilon=0.15$ & & & 39524\\
    & $\epsilon=0.2$ & & & 51191\\
    & $\epsilon=0.25$ & & & 57272\\
    & $\epsilon=0.3$ & & & 60287\\
    \hline
    \multirow{6}{0.7cm}{PGD} & $\epsilon=0.05$ & \multirow{6}{*}{60000\textsuperscript{*}} & \multirow{6}{*}{10000\textsuperscript{**}} & 8671 \\
    & $\epsilon=0.1$ & & & 55432 \\
    & $\epsilon=0.15$ & & & 69604 \\
    & $\epsilon=0.2$ & & & 69818 \\
    & $\epsilon=0.25$ & & & 69823 \\
    & $\epsilon=0.3$ & & & 69823 \\
    \hline
    \multirow{6}{0.7cm}{MIM} & $\epsilon=0.05$ & \multirow{6}{*}{60000\textsuperscript{*}} & \multirow{6}{*}{10000\textsuperscript{**}} & 8679 \\
    & $\epsilon=0.1$  & & & 53150 \\
    & $\epsilon=0.15$ & & & 69402 \\
    & $\epsilon=0.2$  & & & 69822 \\
    & $\epsilon=0.25$ & & & 69823 \\
    & $\epsilon=0.3$  & & & 69825 \\
\end{tabular}
\begin{flushleft}
    \scriptsize
    * saliency maps of MNIST training images \\
    ** saliency maps of MNIST test images
\end{flushleft}
\label{tab:data}
\end{table}

%% file: figures/tab-bench.tex
%!TEX root=appendix.tex

\begin{table}[H]
\caption{Performance benchmark results on detection accuracy, F1 score, and avg AUC of the our detection method. The results are aggregated over multiple class labels.}
\begin{tabular}{ll|c|c|c}
    \multicolumn{2}{c|}{\textbf{Adv. Attack}} & \textbf{Accuracy} & \textbf{F1 Score} & \textbf{Avg. AUC} \\
    \hline \hline
    \multirow{6}{0.7cm}{FGSM} & $\epsilon=0.05$ & 0.8772 & 0.7680 & 0.9299 \\
    & $\epsilon=0.1$ & 0.9233 & 0.9230 & 0.9583 \\
    & $\epsilon=0.15$ & 0.9470 & 0.9543 & 0.9690 \\
    & $\epsilon=0.2$ & 0.9573 & 0.9643 & 0.9733 \\
    & $\epsilon=0.25$ & 0.9644 & 0.9725 & 0.9775 \\
    & $\epsilon=0.3$ & 0.9693 & 0.9747 & 0.9813 \\
    \hline
    \multirow{6}{0.7cm}{PGD} & $\epsilon=0.05$ & 0.9301 & 0.8898 & 0.9723 \\
    & $\epsilon=0.1$ & 0.9839 & 0.9848 & 0.9942 \\
    & $\epsilon=0.15$  & 0.9884 & 0.9896 & 0.9960 \\
    & $\epsilon=0.2$ & 0.9894 & 0.9909 & 0.9961 \\
    & $\epsilon=0.25$ & 0.9898 & 0.9908 & 0.9962 \\
    & $\epsilon=0.3$ & 0.9901 & 0.9912 & 0.9963 \\
    \hline
    \multirow{6}{0.7cm}{MIM} & $\epsilon=0.05$ & 0.9416 & 0.9096 & 0.9799 \\
    & $\epsilon=0.1$ & 0.9839 & 0.9852 & 0.9944 \\
    & $\epsilon=0.15$ & 0.9882 & 0.9899 & 0.9959\\
    & $\epsilon=0.2$ & 0.9897 & 0.9910 & 0.9960 \\
    & $\epsilon=0.25$ & 0.9902 & 0.9915 & 0.9961 \\
    & $\epsilon=0.3$ & 0.9910 & 0.9924 & 0.9965
\end{tabular}
\label{tab:perf}
\end{table}